\documentclass{article}

\usepackage{microtype}
\usepackage{graphicx}
\usepackage{subcaption}
\usepackage{booktabs} 

\usepackage{hyperref}

\usepackage[preprint]{icml2026}

\usepackage{amsmath}
\usepackage{amssymb}
\usepackage{mathtools}
\usepackage{amsthm}

\usepackage[capitalize,noabbrev]{cleveref}

\theoremstyle{plain}

\theoremstyle{definition}

\theoremstyle{remark}

\usepackage[textsize=tiny]{todonotes}

\icmltitlerunning{}

\begin{document}

\twocolumn[
  \icmltitle{SPARK: Self-Play with Asymmetric Reward from Knowledge Graphs}

   \begin{center}
    {\bf Hyobin Park$^{1}$ \quad Taeseop Kim$^{2}$ \quad Dong-Geol Choi$^{2\dagger}$}
 
    $^{1}$ANTLAB, South Korea \quad
    $^{2}$Hanbat National University, South Korea
   
    {\tt hyobinpark03@gmail.com} \quad
    {\tt taeseop.kim@edu.hanbat.ac.kr} \quad
    {\tt dgchoi@hanbat.ac.kr}
   
    $^\dagger$Corresponding author
  \end{center}

  \icmlkeywords{Machine Learning, ICML, Knowledge Graph, Self-Play, Vision-Language Model}

  \vskip 0.3in
]

\printAffiliationsAndNotice{}

\begin{abstract}
Self-play reinforcement learning has shown strong performance in domains with formally verifiable structure, such as mathematics and coding, where both problem generation and reward computation can be grounded in explicit rules. 
Extending this paradigm to scientific literature is more challenging: the relationships among multimodal elements within and across documents are rarely made explicit in text, which makes automatic generation of relational reasoning questions difficult and weakens the reliability of reward signals.
We propose \textbf{SPARK} (\textbf{S}elf-\textbf{P}lay with \textbf{A}symmetric \textbf{R}eward from \textbf{K}nowledge Graphs), a framework that automatically constructs a unified knowledge graph (KG) from multi-document scientific literature and uses it as the structural basis for self-play. KG paths over multimodal nodes serve as a source for generating relational reasoning questions, and structured facts stored in the KG provide a basis for verifiable reward computation. 
A single small vision-language model (sVLM) alternates between Proposer and Solver roles under information asymmetry against a fixed KG, a design that we believe can be naturally extended toward online adaptation in future work.
We evaluate SPARK on public benchmarks and a self-constructed cross-document multi-hop QA dataset. Results show that SPARK consistently outperforms flat-corpus-based self-play baselines, and the performance gap widens as hop count increases, suggesting that KG-structure grounding contributes to relational multi-hop reasoning beyond what unstructured corpus grounding can provide.
\end{abstract}

\section{Introduction}

Scientific progress depends on the accumulation and interconnection of knowledge across the literature. Researchers derive new insights not merely by reading individual papers, but by synthesizing the evolution of concepts, divergences in methodology, and comparisons of experimental results spanning multiple documents. Understanding scientific literature is therefore fundamentally a problem of relational reasoning. Within a single document, figures support specific claims, tables ground experimental measurements, and equations formalize methodology. Across multiple documents, the methodology of one paper connects to the experimental design of another, and results from different works are compared and synthesized. How to train a model to automatically understand this complex relational structure is the central question motivating this work.

\paragraph{The Gap Between Self-Play RL and Unstructured Domains.}
Self-play reinforcement learning is a promising paradigm for improving models without external annotation. Systems such as DeepSeek-R1~\cite{guo2025deepseek} and OpenAI o1~\cite{jaech2024openai} have achieved expert-level performance on mathematics and coding tasks through self-play-based reinforcement learning. The success of this paradigm rests on two conditions. First, formal structure (symbolic equation systems, compilers) enables automatic generation of high-quality problems. Second, that same structure guarantees reliable reward computation that can determine correctness without human involvement. SPICE~\cite{liu2025spice} attempted to extend this paradigm to document corpora, but treating documents as flat text provides no means to represent relationships among multimodal elements or semantic connections across documents, structurally limiting the satisfaction of both conditions.

The relational structure of scientific literature is largely destroyed the moment it is serialized into a linear text sequence. Which claim a figure supports, or how the methodologies of two papers relate, is nowhere explicitly stated in the text. Flat-corpus self-play therefore encounters two fundamental bottlenecks. From the perspective of problem generation, the space of generatable questions is confined to surface-level fact extraction, making automatic generation of relational reasoning questions structurally infeasible. From the perspective of reward computation, when ground-truth answers are scattered across unstructured text, the relational structure grounding those answers remains implicit, making it difficult for any reward scheme to reliably distinguish structurally sound reasoning from plausible-sounding incorrect answers.

\paragraph{A Structural Solution via Knowledge Graphs.}
We argue that knowledge graphs (KGs) provide a principled structural solution to both bottlenecks simultaneously. A KG encodes the multimodal components of a document and the relationships among them as explicit node-edge structures. On this structure, KG paths serve as a source for automatically generating relational reasoning questions (resolving bottleneck~1), and facts stored in the KG provide the basis for verifiable reward computation (resolving bottleneck~2). Furthermore, connecting concept nodes across multiple documents creates cross-document reasoning paths that extend beyond single-document scope, directly incorporating the complex cross-document knowledge synthesis that scientific literature understanding fundamentally requires into the self-play training objective.

\paragraph{SPARK.}
Building on these insights, we propose \textbf{SPARK}. SPARK automatically constructs a three-stage KG from scientific papers via Structural, Reference, and Semantic Relation graphs, and internalizes relational reasoning capabilities through a self-play loop in which a single sVLM alternates between the Proposer role (generating questions conditioned on KG paths) and the Solver role (answering without KG access, under information asymmetry). The reward signal decomposes into three KG-verified components, $R_{\text{answer}}$, $R_{\text{path}}$, and $R_{\text{consistency}}$, directly supervising the structural faithfulness of reasoning paths beyond simple answer matching.

\paragraph{Contributions.}
This paper makes three primary contributions.

\begin{enumerate}

\item \textbf{A KG-Grounded Self-Play Framework.} We propose SPARK, which leverages a fixed KG as an offline knowledge structure to simultaneously satisfy the two conditions for successful self-play by automatically generating high-quality problems and computing verifiable rewards in the scientific literature domain. Through the path faithfulness reward $R_{\text{path}}$, SPARK directly targets multi-hop relational reasoning that flat-corpus self-play cannot supervise.

\item \textbf{A Multimodal Three-Stage KG Construction Pipeline.} For a KG to function as a structural solution, automatically extracting relational structure from raw documents is itself a prerequisite. We present a three-stage pipeline (Structural-Reference-Semantic) that constructs a multimodal KG spanning text, figures, tables, and equations without manual annotation. Cross-Document KG Federation connects concept nodes across papers, enabling cross-document reasoning paths.

\item \textbf{A Cross-Document Multi-Hop QA Benchmark.} The relational multi-hop reasoning capability targeted by SPARK cannot be directly measured by existing benchmarks such as ScienceQA\cite{lu2022scienceqa} or DocVQA\cite{mathew2021docvqa}. We introduce a cross-document QA dataset of 450 questions  derived from 50 arXiv papers.

\end{enumerate}

\begin{figure*}[ht]
  \vskip 0.2in
  \begin{center}
    \centerline{\includegraphics[width=\textwidth]{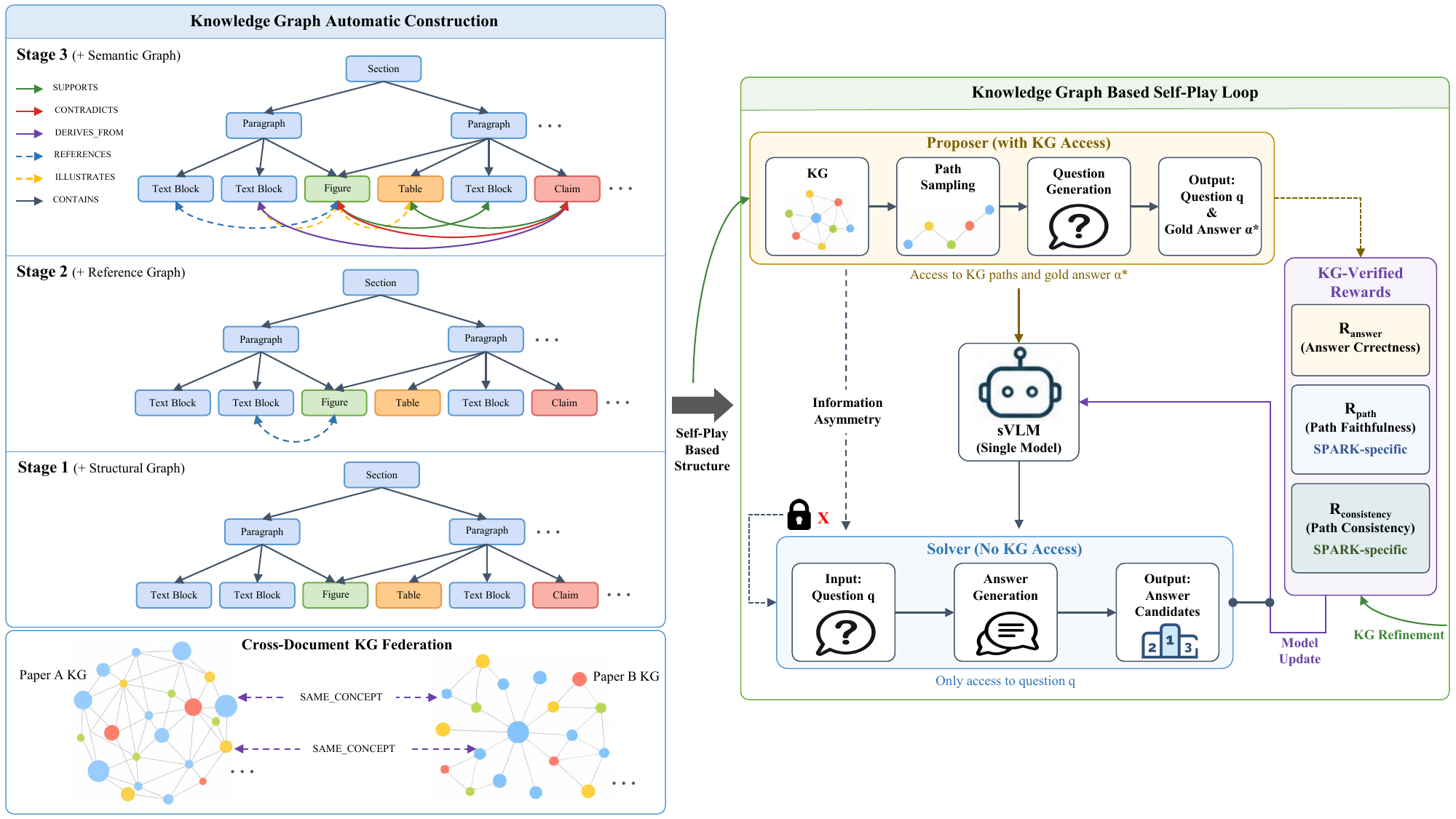}}
    \caption{
      \textbf{Overview of SPARK.} A three-stage pipeline automatically constructs a KG from scientific documents, and Cross-Document Federation connects concept nodes across papers. A single sVLM alternates between Proposer and Solver roles: the Proposer samples reasoning paths from the KG to generate relational questions with gold answers, while the Solver answers without KG access, under information asymmetry. The reward signal decomposes into three KG-verified components, $R_{\text{answer}}$, $R_{\text{path}}$, and $R_{\text{consistency}}$, where $R_{\text{path}}$ and $R_{\text{consistency}}$ are terms unique to SPARK that cannot be defined without the KG.
    }
    \label{overview_img}
  \end{center}
\end{figure*} 

\section{Related Work}
\subsection{Self-Play Reinforcement Learning and Language Models}
Alignment methodologies based on human and AI feedback \cite{ouyang2022, bai2022} have established the standard paradigm for language model post-training. Subsequently, self-play-based approaches \cite{chen2024spin, yuan2024self}, where a single model concurrently serves as both generator and discriminator, have been proposed. However, these methods suffer from an information symmetry bottleneck, where it is difficult to generate independent learning signals since both components share the same prior knowledge. While some studies \cite{wu2025selfplay} have formulated this from a game-theoretic perspective, they primarily focus on alignment rather than factual reasoning. On the other hand, reinforcement learning (RL) based on verifiable rewards \cite{guo2025deepseek} has proven effective in inducing long-chain reasoning. Efforts to extend this to zero-data self-play without external data \cite{zhao2025, huang2025rzero} are also underway. More recently, corpus-based self-play \cite{liu2025spice}, which mines tasks by using unstructured corpora as grounding, has been proposed to alleviate the information symmetry problem. However, unstructured text presents challenges in relational verification and limits the precision of learning signals. Our work inherits this paradigm but replaces the grounding environment with a Knowledge Graph (KG). By utilizing KG paths as a source for automatic relational reasoning task generation and leveraging facts stored in the KG for verifiable reward computation, we simultaneously achieve two success conditions that flat, corpus-based self-play could not structurally satisfy.

\subsection{Knowledge Graph-based Question Answering}

Knowledge Graphs (KGs) provide an explicit multi-step reasoning framework and a verifiable factual foundation through entity-relation triplets, leading to extensive research in their application to question answering (KGQA). Early embedding-based multi-step KGQA methodologies \cite{saxena2020, jiang2023} laid the groundwork, which evolved into approaches where LLMs directly utilize KG paths as reasoning plans \cite{sun2024tog, luo2024rog} and industrial-scale demonstrations of the utility of graph-structured retrieval \cite{edge2024}. On the theoretical front, studies have conceptualized LM reasoning as a weighted aggregation of random walk paths on KGs \cite{wang2024rpa} or improved KG reasoning performance through query-prototype attention \cite{liu2024kf}. Recent works have proposed integrating KG structures directly into the LLM decoding process to fundamentally prevent hallucinations \cite{luo2025gcr} or applying RL-tuned LLM reasoning to relation-centric KGQA \cite{tang2025}. However, the latter lacks an automated task-generation mechanism based on self-play. Our research utilizes the KG for both task generation and reward calculation within a self-play loop. Through an information asymmetry structure—where the Proposer accesses KG paths while the Solver answers without the KG—we directly target cross-document multi-hop reasoning as a primary learning objective.

\subsection{Scientific Document Understanding}

Scientific documents contain complex multimodal constituents such as equations, tables, and figures, along with domain-specific notations, making it difficult to apply general document understanding methods. In terms of parsing and structural extraction, the field has progressed from rule-based and CRF-based methods \cite{lopez2009} to multimodal pre-training \cite{huang2022lmv3}, OCR-free end-to-end transformation \cite{kim2022donut, blecher2023}, and visual document retrieval using vision-language model-based page embeddings \cite{faysse2024}. Regarding language models, since the effectiveness of science-specific pre-training \cite{beltagy2019} was demonstrated, scaling these to large-scale decoder models \cite{taylor2022, lewkowycz2022} has significantly advanced language understanding and generation in the scientific domain. Building upon these document understanding technologies, our work automatically constructs a three-tier KG (Structural, Reference, and Semantic) from multimodal documents—including text, figures, tables, and equations~\cite{seo2025upsampleanything}—without manual annotation. By utilizing this as a grounding environment for self-play, we transform the relational reasoning inherently required by scientific literature into a learnable structural signal.

\section{Method: SPARK}

SPARK consists of two phases. Phase~1 automatically constructs a fixed KG, and Phase~2 uses this KG as the sole structural foundation of a self-play loop to internalize relational reasoning capabilities into a single sVLM.

\subsection{Automatic KG Construction}

\subsubsection{Three-Stage KG Construction Pipeline}

Leveraging the relational structure of scientific documents for self-play requires restoring implicit relationships that are lost during text serialization into explicit graph structure. These relationships exist at multiple distinct levels: the physical layout of a document, the cross-modal connections intended by the author, and the semantic or causal relationships between content units. We address this through a three-stage pipeline, where each stage independently recovers one level of structure.

The KG follows schema $\mathcal{G} = (\mathcal{V}, \mathcal{E})$. The node type set $\mathcal{V} = \{\text{TextBlock, Figure, Table, Equation, Concept, Claim}\}$ is designed to cover the full range of multimodal components in scientific literature. The edge type set $\mathcal{E}$ is designed to span three levels: hierarchical containment (\textsc{Contains}, \textsc{HasCaption}), cross-modal reference (\textsc{References}, \textsc{Illustrates}, \textsc{Quantifies}, \textsc{Defines}), and semantic relations (\textsc{Supports}, \textsc{Contradicts}, \textsc{DerivesFrom}, \textsc{Compares}). This taxonomy corresponds precisely to the three levels of structure recovered by the stages below. Each edge $e$ carries a confidence score $c(e) \in [0, 1]$ that is updated incrementally during self-play.

\paragraph{Stage 1: Structural Graph.}
The physical hierarchy of a document, such as section-paragraph containment and the connections between figures, tables, or equations and their adjacent text, is encoded via rule-based parsing to form the skeletal scaffold of the KG. This stage is deterministic and noise-free, but its fundamental limitation is that the relationships it captures are restricted to physical adjacency. Which claim a figure supports, or which equation underlies a particular experimental result, cannot be recovered at this stage.

\paragraph{Stage 2: Reference Graph.}
Cross-modal connections intended by the author exist independently of physical adjacency. A sentence citing a figure (e.g., ``As shown in Figure~3'') expresses a semantic link between a claim and that figure regardless of where the figure appears in the document flow. We parse such cross-reference expressions via regular expression pattern matching to generate \textsc{References} edges. Through this stage, figures, tables, and equations acquire roles within the KG not merely as visual elements, but as evidence grounding specific claims.

\paragraph{Stage 3: Semantic Relation Graph.}
The first two stages recover the surface structure of documents, but causal, supportive, and contradictory relationships, which represent the core of scientific reasoning, can only be extracted through semantic understanding of content. We first filter candidate node pairs by cosine similarity of their embeddings, then apply a VLM to classify the relationship for high-priority pairs into one of \{\textsc{Illustrates}, \textsc{Supports}, \textsc{Contradicts}, \textsc{DerivesFrom}, \textsc{Compares}\}. Only at this stage do non-trivial multi-hop reasoning paths emerge within the KG, capturing relationships such as which experimental result supports which theoretical claim, or how two methodologies compare.

The KG produced by these three stages simultaneously satisfies both conditions required for self-play. A path $P = (n_0, e_1, n_1, \ldots, e_k, n_k)$ over the KG serves as the source for question generation (\S\ref{sec:proposer}), the content of the terminal node provides the gold answer $a^*$, and the full set of KG-stored facts provides the basis for reward verification (\S\ref{sec:reward}).

\subsubsection{Cross-Document KG Federation}

Constructing a KG within each paper in isolation precludes learning that integrates knowledge across multiple documents. Rather than treating individual documents as independent knowledge units, we connect shared concepts, conflicting claims, and complementary methodologies across documents within a single unified KG, enabling the model to internalize the ability to combine cross-document knowledge through the self-play process.

For any pair of Concept, Claim, or TextBlock nodes belonging to different papers, we insert a \textsc{SameConcept} edge if the cosine similarity of their embeddings exceeds threshold $\tau$. Once a \textsc{SameConcept} edge is inserted, reasoning paths that cross document boundaries exist within the KG. For example, a 3-hop path of the form
\[
\text{Method}_A \xrightarrow{\textsc{SameConcept}} \text{Method}_B \xrightarrow{\textsc{Contradicts}} \text{Result}_B
\]
captures how a methodological difference between two papers leads to a discrepancy in experimental results, and this path becomes the source for Proposer's cross-document question generation.

\begin{table*}[!t]
  \caption{Main performance comparison on public multimodal benchmarks.}
  \label{tab:public}
  \begin{center}
    \begin{small}
      \begin{sc}
        \begin{tabular}{llccc}
          \toprule
          Model & Params & ScienceQA & DocVQA & ChartQA \\
                &        & (Acc$\uparrow$) & (ANLS$\uparrow$) & (Acc$\uparrow$) \\
          \midrule
          InternVL2           & 4B & \textbf{94.39} &89.2  &81.5 \\
          \midrule
          Qwen3-VL (vanilla)  & 4B &90.14 &95.3 &84.6 \\
          \quad +SFT only     & 4B &91.6 &95.6 &86.29 \\
          \quad +SPIN (no KG) & 4B &89.4 &95.57 &83.14 \\
          \midrule
          Ours (KG Self-Play) & 4B &93.0 & \textbf{96.8} & \textbf{89.02} \\
          \bottomrule
        \end{tabular}
      \end{sc}
    \end{small}
  \end{center}
  \vskip -0.1in
\end{table*}

\begin{table*}[t]
  \caption{KG-grounded multi-hop reasoning performance.}
  \label{tab:kg_reasoning}
  \begin{center}
    \begin{small}
      \begin{sc}
        \begin{tabular}{lcccccc}
          \toprule
          Model & 1-hop & 2-hop & 3-hop & Faithfulness & Halluc. Rate \\
           & (Acc$\uparrow$) & (Acc$\uparrow$) & (Acc$\uparrow$) & ($\uparrow$) & ($\downarrow$) \\
          \midrule
          Qwen3-VL-4B (vanilla) &59.33 &54.73 &51.8 &59.45 &12.70 \\
          \quad +SFT only       &59.9 &55.18 &53.25 &60.5 &15.97 \\
          \quad +SPIN (no KG)   &57.27 &54.73 &53.32 &60.43 &12.7 \\
          \midrule
          Ours (4B) & \textbf{63.67} & \textbf{61.24} & \textbf{60.32} & \textbf{66.89} & \textbf{9.09}  \\
          \bottomrule
        \end{tabular}
      \end{sc}
    \end{small}
  \end{center}
  \vskip -0.1in
\end{table*}

\begin{table}[t]
  \caption{Reward component ablation.}
  \label{tab:reward}
  \begin{center}
    \begin{small}
      \begin{sc}
        \begin{tabular}{lcccc}
          \toprule
          Reward & ScienceQA & 2-hop & Path F1 \\
          \midrule
          $R_{a}$ only              & 91.2 & 38.4 & 0.41 \\
          $R_{a}+R_{p}$             & 91.8 & 42.1 & 0.57 \\
          $R_{a}+R_{c}$             & 92.1 & 40.3 & 0.44 \\
          All $(R_{a}+R_{p}+R_{c})$ & \textbf{93.1} & \textbf{44.8} & \textbf{0.63} \\
          \bottomrule
        \end{tabular}
      \end{sc}
    \end{small}
  \end{center}
  \vskip -0.1in
\end{table}

\subsection{KG-Grounded Self-Play Learning}

The self-play loop in SPARK is formalized as a single sVLM $\pi_\theta$ alternating between the Proposer and Solver roles. The learning objective is:
\begin{equation}
J(\theta) = \mathbb{E}_{P \sim \mathcal{G}} \left[ \mathbb{E}_{q,\, a^* \sim \pi_\theta(\cdot \mid P,\, \textsc{Proposer})} \left[ \sum_{i=1}^{G} R_{\mathcal{G}}(\hat{a}_i, a^*, P) \right] \right]
\end{equation}
where $\hat{a}_i \sim \pi_\theta(\cdot \mid q, \textsc{Solver})$, and the KG-verified reward $R_{\mathcal{G}}$ decomposes as:
\begin{multline}
R_{\mathcal{G}}(\hat{a}, a^*, P) = w_a \cdot R_{\text{answer}}(\hat{a}, a^*) + w_p \cdot R_{\text{path}}(P, \mathcal{G}) \\
+ w_c \cdot R_{\text{consistency}}(\hat{a}, P)
\end{multline}
Unlike prior self-play methods that rely solely on $R_{\text{answer}}$, both $R_{\text{path}}$ and $R_{\text{consistency}}$ are terms that cannot be defined without $\mathcal{G}$. The fact that two of the three reward components are verified against the KG is the formal expression of our central claim that the KG serves as the structural backbone of the reward signal. The full procedure is described in Algorithm~\ref{alg:spark}.

\subsubsection{Proposer: KG Path-Conditioned Question Generation}
\label{sec:proposer}

The Proposer's role is to automatically generate relational reasoning questions from reasoning paths over the KG. The key design principle is that question type and difficulty are structurally determined by the edge types and length of the sampled path, so that a systematic curriculum emerges naturally without requiring a separate difficulty estimation step.

\paragraph{Reasoning Path Sampling.}
We perform a weighted random walk over $\mathcal{G}$ to sample a path $P = (n_0, e_1, n_1, \ldots, e_k, n_k)$. Edge sampling weights are proportional to semantic importance and inversely proportional to the per-edge-type accuracy from the previous epoch, so that edge types on which the model consistently underperforms are prioritized in the next epoch, yielding automatic curriculum adjustment. The path length $k$ increases progressively from 1-hop to 3 or more hops as training advances.

\paragraph{Question Generation and Information Asymmetry.}
Given the node contents and edge types of path $P$, the Proposer generates a question $q$ using an edge-type-specific prompt template, and extracts the gold answer $a^*$ from the content of the terminal node. For paths that include Figure or Table nodes, the corresponding image is provided as part of the input, naturally inducing visual-linguistic reasoning questions. The Solver receives only $q$ and has no access to $P$ or $a^*$. This information asymmetry, where the Proposer knows the KG path and gold answer while the Solver does not, is the sole source of learning signal in the self-play loop.

\subsubsection{Solver: Answer Generation under Information Asymmetry}

The Solver generates $G$ candidate answers conditioned solely on question $q$, without access to the KG or the reasoning path. The same model instance $\pi_\theta$ serves as both Proposer and Solver, and updated adapter weights are immediately reflected in both roles after each update step. Because information asymmetry is the sole source of the learning signal, the structural diversity of KG paths directly determines the capacity of the self-play loop to generate continuously novel challenges.

\subsubsection{KG-Verified Reward Function}
\label{sec:reward}

The fundamental motivation for our reward design is that without an explicit relational structure grounding the correct answer, it is structurally difficult for any reward scheme to distinguish genuinely correct reasoning from plausible-sounding incorrect answers. SPARK resolves this through three verification signals that exploit the KG as a structured fact database.

\paragraph{$R_{\text{answer}}$: Semantic Answer Correctness.}
The alignment between the gold answer $a^*$ extracted from the terminal node of the KG path and the generated answer $\hat{a}$ is measured via a hierarchical matching criterion that considers exact containment, numerical equivalence, and keyword overlap. This design rewards semantically equivalent answers regardless of surface-level expression differences, mitigating the synonym misclassification problem inherent to naive string matching.

\paragraph{$R_{\text{path}}$: Path Faithfulness.}
We verify that each consecutive node pair in the reasoning path $P$ is connected by a valid edge in $\mathcal{G}$:
\begin{equation}
R_{\text{path}}(P, \mathcal{G}) = \frac{1}{k} \sum_{i=0}^{k-1} \mathbf{1}\left[(n_i, n_{i+1}) \in \mathcal{E}_{\mathcal{G}}\right]
\end{equation}
This term is unique to SPARK and cannot exist in flat-corpus self-play. It functions as a quality signal measuring whether the Proposer has generated questions grounded in structurally meaningful KG relations.

\paragraph{$R_{\text{consistency}}$: KG Factual Consistency.}
Numerical values and key concepts present in the generated answer are verified against the facts stored in KG nodes along the reasoning path. By treating KG nodes as a structured fact database, this term directly measures whether the answer is grounded in information from the actual documents, structurally penalizing hallucination.

The weights $(w_a, w_p, w_c)$ are linearly annealed according to the curriculum schedule. In early epochs, $w_a$ is set high to establish basic answer generation capability; as training progresses, the weights of $(w_p, w_c)$ are gradually increased to shift the learning objective from simple fact retrieval toward relational reasoning.

\subsubsection{Model Update and KG Refinement}

\paragraph{Model Update.}
$(q, \hat{a})$ pairs are converted into preference data based on $R_{\mathcal{G}}$, ranking candidate answers by their total reward scores. Positive examples (high-reward answers) and negative examples (low-reward answers) are used to update the shared parameters $\theta$ via policy gradient optimization with LoRA \cite{hu2022lora} adapters. Because both the Proposer and Solver roles are served by the same model instance $\pi_\theta$, each parameter update is immediately reflected in both roles, enabling the Proposer and the Solver to co-evolve within a single model.

\paragraph{KG Refinement.}
A key property of SPARK is that the KG does not remain a static structure but is continuously refined to reflect self-play outcomes. \textit{Missing Edge Detection} adds new edges for cases where the Solver achieves high reward on a path that does not exist in the current KG, indicating that the model has implicitly discovered a relationship not captured during KG construction. \textit{Spurious Edge Pruning} decreases the confidence of edges repeatedly implicated in incorrect answers and removes edges whose confidence falls below a threshold. This creates a bidirectional co-evolution cycle in which model training improves KG quality, and the improved KG in turn provides more accurate reward signals.

\section{Experiments}

\subsection{Experimental Setup}

\paragraph{Models and Training.}
We use Qwen3-VL-4B-Instruct \cite{Qwen3-VL} as the base model, fine-tuned with LoRA \cite{hu2022lora} (rank$=16$, $\alpha=32$) for 3 epochs. We compare against four baselines: (1) Qwen3-VL-4B vanilla, (2) +SFT only, (3) +SPIN \cite{chen2024spin}, a flat-corpus self-play baseline without KG reward, and (4) InternVL2-4B \cite{chen2024far}. The +SPIN baseline applies the identical self-play structure as SPARK but removes all KG-based reward components, serving as the critical ablation for isolating the independent contribution of KG-grounded rewards. Detailed training configurations are provided in Appendix~A

\paragraph{Evaluation.}
We evaluate on public benchmarks (ScienceQA \cite{lu2022scienceqa}, DocVQA \cite{mathew2021docvqa}, ChartQA \cite{masry2022chartqa}) and our self-constructed Cross-Document Multi-hop QA dataset (50 arXiv papers, 450 questions). Dataset construction procedures and metric definitions are detailed in Appendix~B

\subsection{Public Benchmark Results}

Table~\ref{tab:public} reports results on public benchmarks. Qwen3-VL-4B vanilla already constitutes a strong baseline, surpassing the larger InternVL2 (8B+) on select metrics such as ScienceQA (90.14 vs.\ 94.39) and ChartQA (84.6 vs.\ 81.5). In this context, the additional gains achieved by SPARK more clearly attribute the improvement to KG-grounded self-play training rather than to base model capacity. The consistent gains of SPARK over +SPIN, under an otherwise identical self-play structure, empirically isolate the pure contribution of the KG reward signal.

\subsection{Cross-Document Multi-Hop QA Results}

Existing public benchmarks are designed around single-document settings and therefore cannot measure the cross-document relational reasoning that SPARK targets. Table~\ref{tab:kg_reasoning} presents results on our self-constructed dataset, providing direct empirical validation of the paper's central contribution.

We highlight two key patterns. First, the accuracy gap between SPARK and all baselines widens monotonically as hop count increases. The vanilla model exhibits a sharp performance drop from 1-hop to 3-hop, whereas SPARK degrades more gradually. This demonstrates that KG-structure grounding specifically enhances relational multi-hop reasoning beyond surface-level fact extraction. Second, the co-improvement of Path F1 alongside accuracy confirms that SPARK not only reaches the correct answer but does so by traversing the intended KG relational path. In contrast, +SPIN shows limited improvement in Path F1, indicating that structured reasoning paths do not emerge without KG grounding.

\subsection{Reward Component Ablation}

Table~\ref{tab:reward} isolates the contribution of each component of $R_{\mathcal{G}}$, providing the direct experimental basis for our central claim that the KG serves as the structural backbone of verifiable reward computation.

Three observations follow. First, using $R_{\text{answer}}$ alone yields performance comparable to SFT/DPO, demonstrating that the structural advantages of self-play do not materialize without KG-based reward signals. Second, adding $R_{\text{path}}$ produces the largest gains in Path F1 and cross-document accuracy, while adding $R_{\text{consistency}}$ contributes relatively more on public benchmarks. The two components thus address complementary axes of reasoning quality, acting synergistically within the full $R_{\mathcal{G}}$. Third, the independent contributions of $R_{\text{path}}$ and $R_{\text{consistency}}$ constitute the direct experimental evidence for the central claim of this paper.

\subsection{Qualitative Analysis}

\begin{figure*}[ht]
  \vskip 0.2in
  \begin{center}
    \centerline{\includegraphics[width=\textwidth]{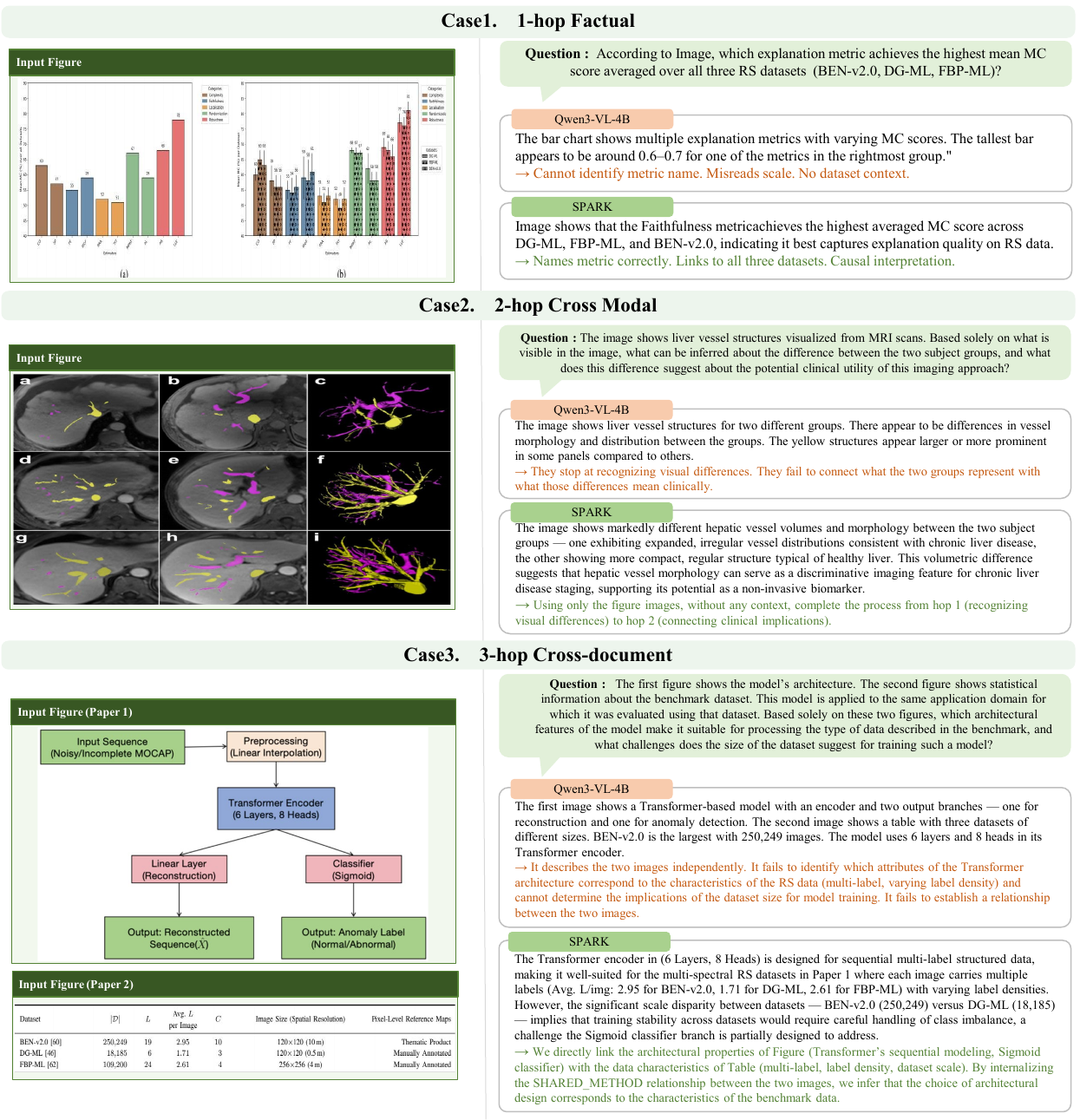}}
    \caption{
      \textbf{Overview of SPARK.} Qualitative comparison of Qwen3-VL-4B vanilla and SPARK across three cases of increasing relational complexity. Both models receive identical inputs at inference time; differences arise solely from model weights. 
    }
    \label{fig:qualitative}
  \end{center}
\end{figure*}

Figure~\ref{fig:qualitative} presents a direct comparison of responses from Qwen3-VL-4B vanilla and SPARK on three representative QA cases across difficulty levels. In all cases, both models receive identical inputs (question and image), and any difference in output quality is attributable solely to model weights.

In \textbf{Case~1} (1-hop, Factual), the vanilla model reads approximate numerical values from the bar chart but fails to connect the metric name to its dataset context. SPARK, by contrast, correctly identifies the Faithfulness metric and provides an interpretation of the average across all three RS datasets.

In \textbf{Case~2} (2-hop, Cross-modal), the vanilla model confines its response to describing visual differences in the image. SPARK completes the 2-hop connection from visual observation to clinical implication (non-invasive biomarker) using the image alone.

In \textbf{Case~3} (3-hop, Cross-document), a Figure and a Table retrieved from two separate papers via KG retrieval are provided as shared input to both models. The vanilla model describes the two images independently without connecting them. SPARK performs cross-document relational reasoning, directly linking properties of the Transformer architecture to characteristics of RS benchmark data.

Across all three cases, the gap in answer quality between the two models widens as hop count increases, consistent with the quantitative results reported in Table~\ref{tab:multihop}.

\clearpage

\section{Conclusion}

We present SPARK, a framework that addresses the fundamental barriers encountered when extending self-play beyond domains with formally verifiable structure. The core insight is that a knowledge graph automatically constructed from scientific documents simultaneously restores the two conditions that self-play requires: a source for automatic generation of relational reasoning problems, and a basis for verifiable reward computation. By treating the KG not as a retrieval index but as the shared foundation for both problem generation and reward verification, SPARK enables a form of self-improvement that flat-corpus methods cannot achieve. Rather than extracting surface-level facts, the model learns to traverse chains of relations that span multimodal document elements.

Through the information asymmetry between a Proposer that constructs questions from KG paths and a Solver that answers without KG access, SPARK continuously generates an ever-expanding space of relational reasoning challenges. As Cross-Document KG Federation broadens the scope of the KG and KG Refinement drives co-evolution between the model and the graph, the difficulty and diversity of the self-generated curriculum scale naturally with the breadth of knowledge encoded in the graph. The model consequently internalizes not merely the facts contained in individual documents, but the relational structure connecting concepts, methodologies, and experimental findings across the entire corpus.

This shift in perspective, treating documents not as collections of text but as structured relational environments for self-play, opens a new direction for training annotation-free scientific reasoning models. The underlying principle is not specific to scientific literature; it extends to any domain in which structured relational knowledge exists implicitly, including legal documents, medical records, and technical specifications. As the body of human knowledge encoded in documents continues to grow, frameworks such as SPARK that can surface relational learning signals from that structure will become increasingly important.

\nocite{langley00}

\bibliography{example_paper}
\bibliographystyle{icml2026}

\newpage
\appendix
\onecolumn
\section{Detailed Training Configuration}
\paragraph{Base Model and Training Framework.}
We use Qwen3-VL-4B-Instruct as the base model. Fine-tuning is performed with LoRA~\citep{hu2022lora}, configured with rank $r=16$, scaling factor $\alpha=32$, and dropout $p=0.05$. LoRA adapters are applied to the query projection ($q_{\text{proj}}$) and value projection ($v_{\text{proj}}$) layers of the attention mechanism. Model parameters are updated via GRPO (Group Relative Policy Optimization), with role-specific advantages computed separately for the Proposer and Solver:
\begin{equation}
A_{\text{Proposer},i} = r_{\text{Proposer},i} - \bar{r}_{\text{Proposer}},
\qquad
A_{\text{Solver},i} = r_{\text{Solver},i} - \bar{r}_{\text{Solver}}
\end{equation}
Although the learning signals for the two roles are computed independently, they are applied to the same shared parameters, enabling the Proposer and Solver to co-evolve within a single model.

\paragraph{Training Hyperparameters.}
Training runs for 30 epochs with 100 questions per epoch. Only $(q, \hat{a})$ pairs whose total reward exceeds the threshold of 0.5 are used as training data. The GRPO group size is set to $G=8$ with a maximum generation length of 256 tokens. We use a learning rate of $2.0 \times 10^{-4}$, a batch size of 2, and a gradient accumulation step of 4, with a maximum of 100 update steps per epoch. The DPO temperature is set to $\beta=0.1$. Mini-batch update mode is supported, applying an immediate policy gradient every 15 examples to enable more frequent intra-epoch updates. A summary of hyperparameters is provided in Table~\ref{tab:hyperparams}.

\paragraph{Reward Weight Schedule.}
The reward weights $(w_a, w_p, w_c)$ are dynamically adjusted according to the curriculum schedule. In early epochs, $w_a$ is set high to prioritize establishing basic answer generation capability. As training progresses, the weights of $(w_p, w_c)$ are gradually increased to shift the learning objective toward structural faithfulness. The default initial values are $(w_a, w_p, w_c) = (0.5, 0.3, 0.2)$.

\paragraph{KG Configuration.}
A cosine similarity threshold of $\tau = 0.75$ is applied for semantic relation graph construction, with a maximum of 10 edges per node. The concept merging threshold for Cross-Document KG Federation is set to $\tau_{\text{cross}} = 0.85$. Edge confidence pruning is applied with threshold $\tau_{\text{prune}} = 0.15$, and the node merging threshold is $\tau_{\text{merge}} = 0.88$. Edge-type sampling weights are assigned according to semantic importance, as listed in Table~\ref{tab:edge_weights}.

\begin{table}[ht]
\caption{Edge-type sampling weights.}
\label{tab:edge_weights}
\vskip 0.15in
\begin{center}
\begin{small}
\begin{tabular}{lc}
\toprule
Edge Type & Sampling Weight \\
\midrule
\textsc{SameConcept}   & 1.00 \\
\textsc{Contradicts}   & 0.95 \\
\textsc{Supports}, \textsc{Illustrates}, \textsc{Compares} & 0.90 \\
\textsc{Defines}, \textsc{DerivesFrom} & 0.85 \\
\textsc{Quantifies}, \textsc{References} & 0.80 \\
\textsc{HasCaption}    & 0.50 \\
\textsc{Contains}      & 0.30 \\
\bottomrule
\end{tabular}
\end{small}
\end{center}
\vskip -0.1in
\end{table}

\paragraph{Embedding Model.}
Node similarity computation and Cross-Document Federation use \texttt{sentence-transformers/all-MiniLM-L6-v2}.

\paragraph{Curriculum Learning.}
The reasoning path length $k$ increases progressively from 1-hop to a maximum of 3 or more hops as training advances. Difficulty is organized into four levels: VQA, Factual, Causal, and Instruction Following. Per-epoch per-edge-type accuracy is tracked, and sampling weights for edge types with below-average accuracy are automatically increased in the subsequent epoch, yielding adaptive curriculum adjustment without manual intervention.

\section{Dataset, Evaluation Metrics}

\subsection{Cross-Document Multi-Hop QA Dataset Construction and Evaluation Metrics}

\paragraph{Data Collection.}
We collect 50 papers from arXiv, balanced across three domains: computer science, physics, and life sciences. Each paper is processed through the SPARK KG pipeline: a three-stage KG is constructed per paper, and Cross-Document KG Federation connects concept nodes across papers.

\paragraph{Question Generation.}
Questions are automatically generated from KG reasoning paths using a large publicly available VLM. The dataset comprises 450 questions in total, with 150 questions per hop level (1-hop, 2-hop, 3-hop), balanced across four question types: Factual, Comparative, Causal, and Synthesis. Final quality is verified through expert review. We plan to expand the dataset to 200 papers and 1,200 questions for the full venue submission.

\paragraph{Evaluation Metrics.}
Accuracy alone is insufficient to measure whether a model reasons through the intended relational path. We therefore report three complementary metrics.

\subparagraph{Accuracy.}
We measure the alignment between the gold answer $a^*$, extracted from the terminal node of the KG reasoning path, and the model output $\hat{a}$ via a hierarchical matching criterion:
\begin{equation}
\text{Acc}(\hat{a}, a^*) = \max \left(
  \mathbf{1}[a^* \subseteq \hat{a}],\;
  \mathbf{1}\!\left[\frac{|\hat{a}_{\text{num}} - a^*_{\text{num}}|}{a^*_{\text{num}}} \leq \epsilon \right],\;
  \frac{|\mathcal{K}(\hat{a}) \cap \mathcal{K}(a^*)|}{|\mathcal{K}(a^*)|}
\right)
\end{equation}
where $\mathcal{K}(\cdot)$ denotes the set of key keywords and $\epsilon = 0.05$ is the numerical tolerance ratio. The three criteria---exact containment, numerical equivalence, and keyword overlap---are evaluated in order, ensuring that semantically equivalent answers expressed in different surface forms receive credit.

\subparagraph{Path F1.}
Path F1 measures the overlap between the set of node pairs comprising the KG reasoning path, $\mathcal{P}_{\text{KG}} = \{(n_i, n_{i+1})\}_{i=0}^{k-1}$, and the set of node pairs referenced by the model, $\mathcal{P}_{\text{model}}$:
\begin{equation}
\text{Path F1} = \frac{2 \cdot |\mathcal{P}_{\text{model}} \cap \mathcal{P}_{\text{KG}}|}{|\mathcal{P}_{\text{model}}| + |\mathcal{P}_{\text{KG}}|}
\end{equation}
A model that achieves high Accuracy but low Path F1 has reached the correct answer by bypassing the intended relational path, indicating an absence of genuine multi-hop relational reasoning.

\subparagraph{Hallucination Rate.}
Hallucination Rate measures the proportion of numerical values and factual claims in the model output that are inconsistent with facts stored in KG nodes. Numerical hallucination (HalNum) and factual hallucination (HalFact) are measured separately and combined as:
\begin{equation}
\text{HalNum} = \frac{1}{|\mathcal{V}|}\sum_{(\hat{v}, v^*) \in \mathcal{V}}
  \mathbf{1}\!\left[\frac{|\hat{v} - v^*|}{|v^*|} > \tau\right],
\qquad
\text{Hallucination Rate} = \frac{1}{2}\left(\text{HalNum} + \text{HalFact}\right)
\end{equation}
where $\tau = 0.05$ is the numerical error tolerance threshold. Because this metric is computed automatically against quantitative facts stored in Table and Equation nodes, it provides an objective measurement that does not rely on the subjectivity of LLM-as-Judge evaluation.

Taken together, the three metrics assess whether a model arrives at the correct answer, via the correct reasoning path, and on the basis of correct evidence.

\end{document}